\documentclass[pmlr]{jmlr}

\usepackage{longtable}

\usepackage{booktabs}
\usepackage[]{algorithm2e}
\usepackage[load-configurations=version-1]{siunitx} 

\theorembodyfont{\upshape}
\theoremheaderfont{\scshape}
\theorempostheader{:}
\theoremsep{\newline}

\jmlrvolume{1}
\jmlryear{2019}
\jmlrworkshop{NeurIPS2019 Disentanglement Challenge}

\title[Disentanglement with Hyperspherical $\Delta$VAE]{Disentanglement with Hyperspherical Latent Spaces using Diffusion Variational Autoencoders}

  \author{\Name{Luis Armando P\'{e}rez Rey} \Email{l.a.perez.rey@tue.nl}\\
   \addr Eindhoven University of Technology. Eindhoven, The Netherlands. \\
   \addr Prosus AI, Prosus. Amsterdam, The Netherlands.
   }
   
\begin{document}

\maketitle

\begin{abstract}
A disentangled representation of a data set should be capable of recovering the underlying factors that generated it. One question that arises is whether using Euclidean space for latent variable models can produce a disentangled representation when the underlying generating factors have a certain geometrical structure. Take for example the images of a car seen from different angles. The angle has a periodic structure but a 1-dimensional representation would fail to capture this topology. How can we address this problem? The submissions presented for the first stage of the  NeurIPS2019 Disentanglement Challenge consist of a Diffusion Variational Autoencoder ($\Delta$VAE) with a hyperspherical latent space which can for example recover periodic true factors. The training of the $\Delta$VAE is enhanced by incorporating a modified version of the Evidence Lower Bound (ELBO) for tailoring the encoding capacity of the posterior approximate.

\end{abstract}

\section{Introduction}
\label{sec:intro}
Variational Autoencoders (VAEs) proposed by \cite{Kingma2014} are an unsupervised learning method that can estimate the underlying generative model that produced a data set in terms of the so-called latent variables. In the context of VAEs, a disentangled representation is obtained when the latent variables represent the true independent underlying factors, which usually have a semantic meaning, that generated the data set. \\

VAEs assume that a data set $\mathcal{X} = \{x_i\}_{i = 1}^N$ consists of $N$ independent and identically distributed data points belonging to a set $X$. A set $Z$ of unobserved latent variables is proposed and the main goal is to maximize the log-likelihood via variational inference using an approximate to the posterior distribution $Q_{Z|x}^{(a)}$ and a decoding distribution $P_{X|z}^{(b)}$ with parameters $a,b$ calculated by neural networks. A prior distribution $P_Z$ is selected before training such that the training of the VAE is carried out by maximizing for each data point the Evidence Lower Bound (ELBO) w.r.t. the neural network weights that calculate $a, b$ given by 
\begin{equation}
    \mathcal{L}(x,a,b) = \mathbb{E}_{z\sim Q^{(a)}_{Z|x}}\left[\log{p_{X|z}^{(b)}(x)}\right]-\mathrm{KL}\left(Q^{(a)}_{Z|x}||P_Z\right)
\end{equation}

To accomplish the disentanglement of latent variables \cite{Higgins2016} proposed to weight the contribution of both terms in the ELBO by using a parameter $\beta\in \mathbb{R}^+$ to change the capacity of encoding of the posterior distribution. The idea of changing the capacity of the encoding distribution was further explored in \cite{Burgess2018} where the Kullback-Leibler divergence term is pushed towards a certain value $C\in \mathbb{R}^+$ in each training step. The combination of both approaches led to a to the following training objective to be maximized,
\begin{equation}\label{eq:ELBO}
    \mathcal{L}(x, a, b) = \mathbb{E}_{z\sim Q^{(a)}_{Z|x}}\left[\log{p_{X|z}^{(b)}(x)}\right]-\beta\left|\mathrm{KL}\left(Q^{(a)}_{Z|x}||P_Z\right)-C\right|.
\end{equation}
The value of $\beta$ is fixed before training and $C$ is increased linearly each epoch of training from a minimum value $C_{min}$ to $C_{max}$. We refer to this procedure as capacity annealing. \\

In some cases the underlying factors that generated a data set have a certain geometrical/topological structure that cannot be captured with the traditional Euclidean latent variables as has been mentioned in \cite{Falorsi2018} and in \cite{Davidson2018}. This problem is referred to as \textit{manifold mismatch}.\\

For the NeurIPS2019 Disentanglement challenge, datasets for local evaluation are provided based on the paper by \cite{Locatello2018}. It is important to note that in such datasets there is at least one underlying factor that has a periodic structure. Take for example the Cars3D dataset consisting of images of cars. In particular, one factor of variation is the azimuthal angle of rotation of the car. The geometrical structure of this factor is circular and thus it is better represented with a periodical latent variable.\\  

The Diffusion Variational Autoencoders $\Delta$VAE presented by \cite{Rey2019} provide a versatile method that can be used to implement arbitrary closed manifolds for a latent space, in particular, hyperspheres. 

\section{Method Overview}

We propose the use of a $\Delta$VAE with hyperspherical latent space coupled with the capacity annealing procedure from Equation \ref{eq:ELBO}. In \cite{Davidson2018} has described that for high dimensional latent spaces, the vanilla VAE from \cite{Kingma2014} behaves similarly to the VAE with a high dimensional hyperspherical latent space. Thus, we have chosen to use a high dimensional hyperspherical latent space of dimension $d$, i.e. $Z = S^d$ since it can provide better representations for periodical latent variables while still maintaining the properties of the vanilla implementation. \\

\section{Method Description}
The Diffusion VAE from \cite{Rey2019} with hyperspherical latent space consists of the following elements: 
\begin{itemize}
	\setlength{\itemsep}{0em}
    \item Hyperspherical latent space embedded in Euclidean latent space $Z = S^{d}\subseteq\mathbb{R}^{d+1}$
	\item Uniform prior $P_Z$ over the hypersphere.
	\item  Posterior distribution $Q_Z^{ \mu_Z, t}$ from a family of solutions to the heat equation over the hypersphere parameterized by location $\mu_Z\in S^{d+1}$ and scale $t \in \mathbb{R}^+$. 
	\item  Decoder distribution $\mathbb{P}_X^{ \mu_X, \sigma_X}$ from a family of normal distributions
	parametrized by location $\mu_X\in X$ (covariance is chosen to be the identity). 
	\item Neural networks to calculate parameters $\boldsymbol{\mu_Z}:X\mapsto S^d$,$\boldsymbol{t}: X\mapsto \mathbb{R}^+$, $\boldsymbol{\mu_X}:S^d\mapsto X$. The encoding neural network $\boldsymbol{\mu_Z}$ is a composition of a multi layer perceptron into $\mathbb{R}^{d+1}$ with a projection function $\boldsymbol{P}$ into the hypersphere. 
	\item Projection map corresponds to $\boldsymbol{P}:\mathbb{R}^{d+1}\mapsto S^{d}$ such that $\boldsymbol{P}(x) = x/\|x\|_2$.
\end{itemize}

During training, there are two key procedures that need to be taken into account: the \textit{reparameterization trick} for sampling the posterior approximate in order to calculate the first term of Equation \ref{eq:ELBO} and the calculation of the Kullback-Leibler divergence between the posterior approximate and the uniform prior for the second term of Equation \ref{eq:ELBO}. 
\paragraph{Reparameterization trick}
In order to approximate the first term of the ELBO, \cite{Kingma2014} proposed the reparameterization trick. In the hypersphere the procedure for sampling $z\sim Q_{Z|x}^{(\mu_Z,t)}$ described in \cite{Rey2019} was implemented. It consists of a random walk of $L$ steps over the hypersphere which approximates to the transition kernel of the Brownian motion over the manifold. 

\begin{algorithm}
\caption{Sampling of $z\sim Q_{Z|x}^{(\mu_Z,t)}$}\label{alg:moore}
Given a data point $x\in \mathcal{X}$ in the data set.
\begin{enumerate*}
  \item Calculate the parameters for the posterior distribution with the corresponding neural networks $t = \boldsymbol{t}(x)$ and $z^{(0)} = \boldsymbol{\mu_Z}(x)$
  \item Repeat for $l \in \{0,1,2,\ldots, L-1\}$ steps
  \begin{itemize}
    \item Sample an auxiliary variable $\epsilon\sim \mathcal{N}(0,I)$ from a $d+1$ dimensional standard normal distribution.
    \item Calculate the $l+1$ step in the random walk $z^{(l+1)} = \boldsymbol{P}\left(z^{(l)}+\epsilon \, t\right)$
  \end{itemize}
 \item The final sampled latent variable $z\sim Q_{Z|x}^{(\mu_Z,t)}$ corresponds to $z = z^{(L)}$
\end{enumerate*}
\end{algorithm}
The sampled latent variables $z$ is then used to estimate the first term of the ELBO and is passed to the decoding neural network.

\paragraph{Kullback-Leibler Divergence}
The Kullback-Leibler divergence between the prior and the posterior is approximated using the formula in \cite{Rey2019} where $\mathrm{Vol}(S^d)$ corresponds to the volume of the hypersphere and is given by

\begin{equation}
\mathrm{KL} \left(Q_Z^{(\mu_Z,t)} \| P_Z \right)\approx - \frac{d}{2} \log(2 \pi t) - \frac{d}{2} + \log (\mathrm{Vol}(S^d)) + \frac{1}{4} d(d-1)\, t.
\end{equation}

\subsection{Hyperparameter Selection}
The hyperparameter values were chosen based on basic implementations described in the corresponding papers: $\beta$ from \cite{Higgins2016}, capacity annealing \cite{Locatello2018} and Diffusion VAE \cite{Rey2019}. The exact values used are presented in the Appendix A.

\bibliography{references}

\begin{thebibliography}{7}
\providecommand{\natexlab}[1]{#1}
\providecommand{\url}[1]{\texttt{#1}}
\expandafter\ifx\csname urlstyle\endcsname\relax
  \providecommand{\doi}[1]{doi: #1}\else
  \providecommand{\doi}{doi: \begingroup \urlstyle{rm}\Url}\fi

\bibitem[Burgess et~al.(2018)Burgess, Higgins, Pal, Matthey, Watters,
  Desjardins, and Lerchner]{Burgess2018}
Christopher~P. Burgess, Irina Higgins, Arka Pal, Loic Matthey, Nick Watters,
  Guillaume Desjardins, and Alexander Lerchner.
\newblock {Understanding disentangling in {\$}{\textbackslash}beta{\$}-VAE}.
\newblock \penalty0 (Nips), 2018.
\newblock URL \url{http://arxiv.org/abs/1804.03599}.

\bibitem[Davidson et~al.(2018)Davidson, Falorsi, De~Cao, Kipf, and
  Tomczak]{Davidson2018}
Tim~R. Davidson, Luca Falorsi, Nicola De~Cao, Thomas Kipf, and Jakub~M.
  Tomczak.
\newblock {Hyperspherical Variational Auto-Encoders}.
\newblock 2018.
\newblock URL \url{http://arxiv.org/abs/1804.00891}.

\bibitem[Falorsi et~al.(2018)Falorsi, de~Haan, Davidson, De~Cao, Weiler,
  Forr{\'{e}}, and Cohen]{Falorsi2018}
Luca Falorsi, Pim de~Haan, Tim~R. Davidson, Nicola De~Cao, Maurice Weiler,
  Patrick Forr{\'{e}}, and Taco~S. Cohen.
\newblock {Explorations in Homeomorphic Variational Auto-Encoding}.
\newblock 7 2018.
\newblock URL \url{http://arxiv.org/abs/1807.04689}.

\bibitem[Higgins et~al.(2016)Higgins, Matthey, Pal, Burgess, Glorot, Botvinick,
  Mohamed, and Lerchner]{Higgins2016}
Irina Higgins, Loic Matthey, Arka Pal, Christopher Burgess, Xavier Glorot,
  Matthew Botvinick, Shakir Mohamed, and Alexander Lerchner.
\newblock {B-Vae: Learning Basic Visual Concepts With a Constrained Variational
  Framework}.
\newblock \emph{Iclr 2017}, \penalty0 (July):\penalty0 1--13, 2016.
\newblock ISSN 1078-0874.
\newblock \doi{10.1177/1078087408328050}.

\bibitem[Kingma and Welling(2014)]{Kingma2014}
Diederik~P Kingma and Max Welling.
\newblock {Auto-Encoding Variational Bayes}.
\newblock In \emph{International Conference on Learning Representations
  (ICLR)}, 2014.
\newblock ISBN 1312.6114v10.
\newblock \doi{10.1051/0004-6361/201527329}.

\bibitem[Locatello et~al.(2018)Locatello, Bauer, Lucic, Gelly, Sch{\"{o}}lkopf,
  and Bachem]{Locatello2018}
Francesco Locatello, Stefan Bauer, Mario Lucic, Sylvain Gelly, Bernhard
  Sch{\"{o}}lkopf, and Olivier Bachem.
\newblock {Challenging Common Assumptions in the Unsupervised Learning of
  Disentangled Representations}.
\newblock pages 1--33, 2018.
\newblock URL \url{http://arxiv.org/abs/1811.12359}.

\bibitem[P{\'{e}}rez~Rey et~al.(2019)P{\'{e}}rez~Rey, Menkovski, and
  Portegies]{Rey2019}
Luis~A. P{\'{e}}rez~Rey, Vlado Menkovski, and Jacobus~W. Portegies.
\newblock {Diffusion Variational Autoencoders}.
\newblock 2019.
\newblock URL \url{http://arxiv.org/abs/1901.08991}.

\end{thebibliography}

\appendix

\section{Hyperparameter Selection}\label{apd:first}
The hyperparameters used for the submissions presented at the NeurIPS2019 challenge are summarized in the following table. Multiple values correspond to different combinations tested for submission to the AICrowd submission platform. 
\begin{table}[htbp]
\floatconts
  {tab:example}%
  {\caption{Hyperparameter Table}}%
  {\begin{tabular}{lll}
  \bfseries Hyperparameter & \bfseries Values &\bfseries Description\\
  d & 10, 20 & Dimensionality of the latent space\\
  $\beta$ & 1, 2.5, 10 & Strength of the capacity annealing\\
  $L$ & 5 & Length of the random walk \\
  $C_{min}$ & 0 & Starting capacity value\\
  $C_{max}$ & 15 & Final capacity value
  \end{tabular}}
\end{table}

\end{document}